\def\year{2019}\relax
\documentclass[letterpaper]{article} 
\usepackage{aaai19}  
\usepackage{times}  
\usepackage{helvet} 
\usepackage{courier}  
\usepackage[hyphens]{url}  
\usepackage{graphicx} 
\def\UrlFont{\rm}  
\usepackage{graphicx}  
\usepackage{amsmath}

\usepackage[normalem]{ulem}
\usepackage{multirow}
\usepackage{booktabs}

\def\etal{\emph{et al.}}
\usepackage{algorithm}
\usepackage[noend]{algpseudocode}
\usepackage{amsfonts}
\usepackage{color}

\usepackage[switch]{lineno}

\title{Mix Dimension in Poincar\'{e} Geometry for 3D Skeleton-based Action Recognition}
\author{Wei Peng\textsuperscript{1}, Jingang Shi\textsuperscript{2}, Zhaoqiang Xia\textsuperscript{3}, Guoying Zhao\textsuperscript{1,}\thanks{Corresponding author~~ This work is accepted by ACM MM2020.}\\
\textsuperscript{1}Center for Machine Vision and Signal Analysis, University of Oulu, Finland\\
\textsuperscript{2}School of Software Engineering, Xi'an Jiaotong University, Xi'an, China \\
\textsuperscript{3}Northwestern Polytechnical University, Xi'an, China;\\
}

\begin{document}

\maketitle

\begin{abstract}
Graph Convolutional Networks (GCNs) have already demonstrated their powerful ability to model the irregular data, e.g., skeletal data in human action recognition, providing an exciting new way to fuse rich structural information for nodes residing in different parts of a graph. In human action recognition, current works introduce a dynamic graph generation mechanism to better capture the underlying semantic skeleton connections and thus improves the performance. In this paper, we provide an orthogonal way to explore the underlying connections. Instead of introducing an expensive dynamic graph generation paradigm, we build a more efficient GCN on a Riemann manifold, which we think is a more suitable space to model the graph data, to make the extracted representations fit the embedding matrix. Specifically, we present a novel spatial-temporal GCN (ST-GCN) architecture which is defined via the Poincar\'e geometry such that it is able to better model the latent anatomy of the structure data. To further explore the optimal projection dimension in the Riemann space, we mix different dimensions on the manifold and provide an efficient way to explore the dimension for each ST-GCN layer. With the final resulted architecture, we evaluate our method on two current largest scale 3D datasets, i.e., NTU RGB+D and NTU RGB+D 120. The comparison results show that the model could achieve a superior performance under any given evaluation metrics with only 40\% model size when compared with the previous best GCN method, which proves the effectiveness of our model.
\end{abstract}

\section{Introduction}

Human action recognition is one of the most important topics in computer vision studies. It could contribute to many potential applications such as human behavior analysis, video understanding, and virtual reality. In general, several different kinds of modalities, e.g., appearance, depth, optical flow, and skeletal data, are utilized in the action recognize tasks. Recently, skeleton-based human action recognition has attracted considerable attention, since the compact skeleton data make the models more efficient and more robust to the variations of viewpoints and environments. 

In this paper, we focus on the problem of 3D skeleton-based action recognition and expect to provide a more robust neural network for this task. Recently, graph convolutional networks (GCNs)~\cite{defferrard2016convolutional,kipf2016semi}, which have been designed well to preserve the natural topology structure of the skeleton, followed by a temporal convolutional network, i.e., ST-GCNs, have been successfully adopted in skeleton-based action recognition~\cite{yan2018stgan,peng2020learning,shi2019two}. Yan~\etal ~first proposed spatial-temporal graph convolutional network~\cite{yan2018stgan} for this task, which decouples the neural architecture into a GCN to capture the spatial information and an 1D convolutional filter to model the dynamic information. At the GCN part, most current ST-GCN approaches provide a pre-defined graph embedding matrix to encode the skeleton topology. This matrix and the skeleton sequence data together are fed into the ST-GCNs to extract high-level representations. However, as mentioned in work~\cite{peng2020learning}, a fixed graph embedding matrix will introduce the constraints into the feature learning progress, which may be harmful for the representation at higher layers, leading to a negative influence to final classification. Therefore, works in ~\cite{shi2019two,peng2020learning} present either a global or a layer-wise dynamic graph generation paradigm to break the learning constraint. Experiments~\cite{shi2019two,peng2020learning} prove that the the dynamic graph generation mechanism could further improve the performance for this task. 

This paper aims to deal with the skeleton-based human action recognition tasks from another perspective. Instead of providing a dynamic graph embedding, we turn to explore a better modelling space for the skeleton graph sequences. Despite the success of current feature representation with deep neural networks in Euclidean space, graph data is proved to exhibit a highly non-Euclidean latent anatomy~\cite{bronstein2017geometric}. However, as far as we know, all the previous ST-GCNs~\cite{yan2018stgan,shi2019two,peng2020learning} are defined in the Euclidean space, which may be not the optimal choice for modelling the hierarchical graph data. We argue that neural network operations directly defined in a data-related space, e.g., Hyperbolic manifold~\cite{benedetti2012lectures}, can benefit the learning processing. 

To this effect, in this paper, we present a spatial-temporal graph convolutional networks on a particular model of hyperbolic geometry, i.e., the Poincaré model~\cite{benedetti2012lectures}. Hyperbolic geometry, which is a non-Euclidean geometry with a constant negative Gaussian curvature, has recently increasingly gained momentum in the context of deep neural networks~\cite{nickel2017poincare,tifrea2018poincar} due to their high efficient capacity and tree-likeliness properties. Building a ST-GCN on hyperbolic geometry could benefit from the hyperbolic distance since the distance between irrelevant samples will be exponentially greater than the distance between similar samples. Our method is orthogonal to the dynamic graph generation ones. Instead of generating a dynamic graph embedding by computing the node embedding similarity, we explore a more reasonable manifold projection such that the projected feature is more suitable for the given embedding matrix. The relationship between samples represented in the hyperbolic space can emphasize similar samples and suppress irrelevant samples. Besides, our method is also more general for graph sequence data since they naturally lie in a non-Euclidean space.

However, adopting deep neural networks with the non-Euclidean settings is challenging since the non-trivial of the principled generalizations of basic operations such as convolutions. Inspired by work~\cite{gulcehre2018hyperbolic}, we get help from the projection between the hyperbolic space and the tangent space. Since there is a bijection between them, the convolutional operations will be conducted on the tangent spaces and then the extracted features will be projected back as a trajectory on the manifold. In this way, we can get the graph embedding on the hyperbolic space by projecting the feature back to the manifold. To further explore the optimal projection dimension in this non-Euclidean space, we mix different dimensions in the hyperbolic space and provide an efficient way to explore the dimension for each graph neural network layer. Finally, with the resulted ST-GCN, we evaluate our method on two currently most challenging 3D skeleton-based datasets, i.e., NTU RGB+D~\cite{shahroudy2016ntu} and NTU RGB+D 120~\cite{liu2019ntu}. Compared with various state-of-the-art approaches,our method gets the best results under any given evaluation protocols. Moreover, in terms of the model size, current best model~\cite{peng2020learning} is even 2.5 times bigger than our model, which proves the effectiveness of our method.
Our contribution can be summarized as follows:
\begin{itemize}
    \item We present a novel spatial temporal graph convolutional network via Poincar\'e geometry, which gives a brand-new ST-GCN to model the graph sequences on the Riemann manifold.
    \item The approach learns a multidimensional structural embedding for each graph based on a Poincar\'e model. To provide a more distinguished representation, an efficient way is provided to explore a better projection space by mixing the dimensions on the Poincar\'e model.
    \item To evaluate the effectiveness of our method, comprehensive experiments are conducted on two current most challenging 3D skeleton-based action recognition tasks. Results shows that our model obtains the best classification accuracy under any given metrics with an efficient fashion.
\end{itemize}

The rest of this paper is organized as follows. Section~\ref{sec:relate} reviews related approaches and discusses their relationships to the present work. Section~\ref{sec:method} gives a detailed description about the methodology and the corresponding neural architecture. We conduct the experiments in Section~\ref{sec:exp} and report the results obtained on the various datasets. In this section, experiments include an ablation experiment performed on the NTU RGB+D dataset and the performance comparisons with the state-of-the-art approaches. Finally, a conclusion is drawn in Section~\ref{sec:conclusion}.

\section{Related Works}\label{sec:relate}

\subsection{Graph Convolutional Networks (GCNs)}

Generalizing convolutional neural networks from regular data, e.g., images, to irregular data, e.g.,graph data, has been an active topic in recent years. As one of the most successful representative, Graph Convolutional Networks (GCNs)~\cite{monti2017geometric,velivckovic2018graph,kipf2016semi,defferrard2016convolutional}, including spatial-temporal graph convolutional networks (ST-GCNs), increasingly attract attention and get promising results in many research fields. 

Mainly, according to how the graph convolutions are defined, the GCNs can be divided into two categories, i.e., the spectral-domain and the spatial-domain methods. The spectral one converts graph data into its spectrum and applies a filter in spectral domain. There are many representative works in this stream~\cite{kipf2016semi,defferrard2016convolutional}, and it could deal with the whole graph at the one time. Meanwhile, it is time-consuming especially for large graphs. Another limitation is that the spectral construction is limited to a single domain since the spectral filter coefficients are basis dependent. If one could construct compatible orthogonal bases across different domains, this problem can be solved. However, such construction requires expert knowledge of the correspondence between domains, which is extremely difficult in most cases. On the contrary, spatial based methods directly design convolution operation in spatial domain~\cite{monti2017geometric,velivckovic2018graph}, which resembles the traditional convolutional filter as it is for images. The spatial-domain methods are more scalable to large graphs as they directly perform the convolution in the graph domain via information aggregation. However, the disadvantage is that it is difficult to model the global structure.  To model the dynamic information for graph sequences, e.g., skeletal clips for action recognition, many ST-GCNs~\cite{yan2018stgan,shi2019two,peng2020learning} are proposed. However, as far as our knowledge, even the skeletal data lies in a Non-Euclidean space, all of the ST-GCNs are defined in the Euclidean space. Instead, in this paper, we propose a brand-new GCN model, which models the human action  in a Non-Euclidean space. We notice that there are also works~\cite{nickel2017poincare,ganea2018hyperbolic,liu2019hyperbolic} define neural networks in Riemannian space, but they are either based on conventional forward networks or just design transforming and aggregating functions for a network with only two or three layers. Instead, we provide a deep spatial temporal graph convolutional network to deal with dynamic graph sequences.
\vspace{-1em}
\subsection{Human Action Recognition}
Action recognition~\cite{yan2018stgan,peng2019video,peng2020learning} is one of the most important areas in both industry and academia. We can find numbers of previous action recognition works based on RGB images or videos as the RGB data is available everywhere in real life. However, one of their disadvantage is that the learnt representations are prone to being distracted since entire areas of video frames are exploited to learn the representations. 

Currently, the skeleton data is more easy to access and become more popular for this task. Compared with RGB based action recognition, skeleton based ones are more compact and more robust to the complicated and changing background. The works on this topic mainly follows three streams: 1) Hand-crafted features, in which works leverage the dynamics of joint motion by using handcrafted features, including utilizing LOP feature to overcome intra-class variance issue~\cite{wang2012mining}, building histograms of 3D joint locations ~\cite{xia2012view}, and modelling 3D geometric relationships in a Lie group~\cite{vemulapalli2014human}. These methods require much expert knowledge. 2) Conventional deep learning approaches, which provide an automatic feature learning strategy and have become the mainstream methods. Work~\cite{kim2017interpretable} rearranges the graph-structure skeleton data and models it as a pseudo-image based on the manually designed transformation rules, such that the constructed grid data could directly benefited from CNNs. Since the input is time series data, there are also many attempts to utilize RNN and LSTM to model the dynamic information. Representative works include~\cite{du2015hierarchical,shahroudy2016ntu,song2017end,zhang2017view,si2018skeleton}, where they model the skeleton sequences by either extending RNN to spatio-temporal domains, dividing human skeleton into parts or providing a fully connected deep LSTM for this task. Nevertheless,  the performance is hard to be further improved since the physical structure and topology of the graph data is not well considered. 3) Methods based on spatio-temporal graph convolutional networks, which are more popular and suitable for this task. Intuitively, skeleton-features can be represented as a graph structure since their components are homeomorphic. Therefore, joints and bones in skeletons can be defined as the vertices and connections of the graph. Yan~\etal ~developed a spatio-temporal graph convolutional network (ST-GCN)~\cite{yan2018stgan} to model the skeleton data as the graph structure, which leverages the powerful ability to model the irregular data and thus achieves better performance than previous methods. ST-GCN becomes a general framework to deal with the skeleton-based action recognition task. Based on this, work in~\cite{shi2019two} explores an global adaptive embedding matrix generation method, which further improves the performance. Peng~\etal ~introduced the neural architecture search and automatically designed a GCN~\cite{peng2020learning} architecture for this task. The work~\cite{peng2020learning} gets the current best result. Our work is also based on ST-GCN, but we provide a much more efficient way to model the graph sequences thus even no need to construct a dynamic graph.
\vspace{-1em}
\subsection{Neural Networks on Riemannian Manifold}

Popular deep learning methods typically embed  data into a low dimensional Euclidean vector space using a strategy such as functional similarity to capture semantic representation. This is straightforward since our intuition for the real-life world is highly related to the Euclidean space. However, in many fields, e.g.,  genomics, social networks, and skeleton-based action recognition, the latent anatomy of the data is well defined by non-Euclidean spaces such as Riemannian manifolds. Most of previous deep neural networks are directly applied to such data and also get promising performance due to the powerful ability of deep learning to capture patterns from data.

However, the capability of an optimal modelling space should not only reduce the computational burden but also further improve the task performances. Recently, embedding hierarchical data in Riemannian space has achieved attractive results and gained popularity in deep learning. For instance, by constructing in Riemannian space, Mathieu \etal ~proposed a Poincar\'e variational Auto-Encoders~\cite{mathieu2019continuous} and shows a better generalisation for the hierarchical structures. Cho \etal provided a Riemannian approach to batch normalization\cite{cho2017riemannian} and also achieved superior performance. Here we focus on Hyperbolic geometry, which is a non-Euclidean geometry with a constant negative Gaussian curvature. One significant intrinsic property in this geometry is the exponential growth. There have already been some attempts on designing neural networks in this space. Specifically, Nickel and Kiela~\cite{nickel2017poincare} reported pioneering research on learning representation in hyperbolic spaces. Then, work in~\cite{ganea2018hyperbolic} introduced Hyperbolic Neural Networks, connecting hyperbolic geometry with deep learning. After that, works also provide hyperbolic analogues of conventional operations, in which several other algorithms have been developed, such as Poincar\'e GloVe~\cite{tifrea2018poincar} and Hyperbolic Attention Networks~\cite{gulcehre2018hyperbolic}. We also find that works~\cite{liu2019hyperbolic,chami2019hyperbolic} construct graph neural networks using hyperbolic geometry, which are similar with our work. However, our model is different from these works since we are dealing with the dynamic graph sequences while they just focus on static graph. Besides, we provide an efficient way to explore the influence of the projection dimensions for the network while none of them touch this problem.

\section{METHODOLOGY}\label{sec:method}

\begin{figure*}
    \centering
    \includegraphics[width=0.9\textwidth]{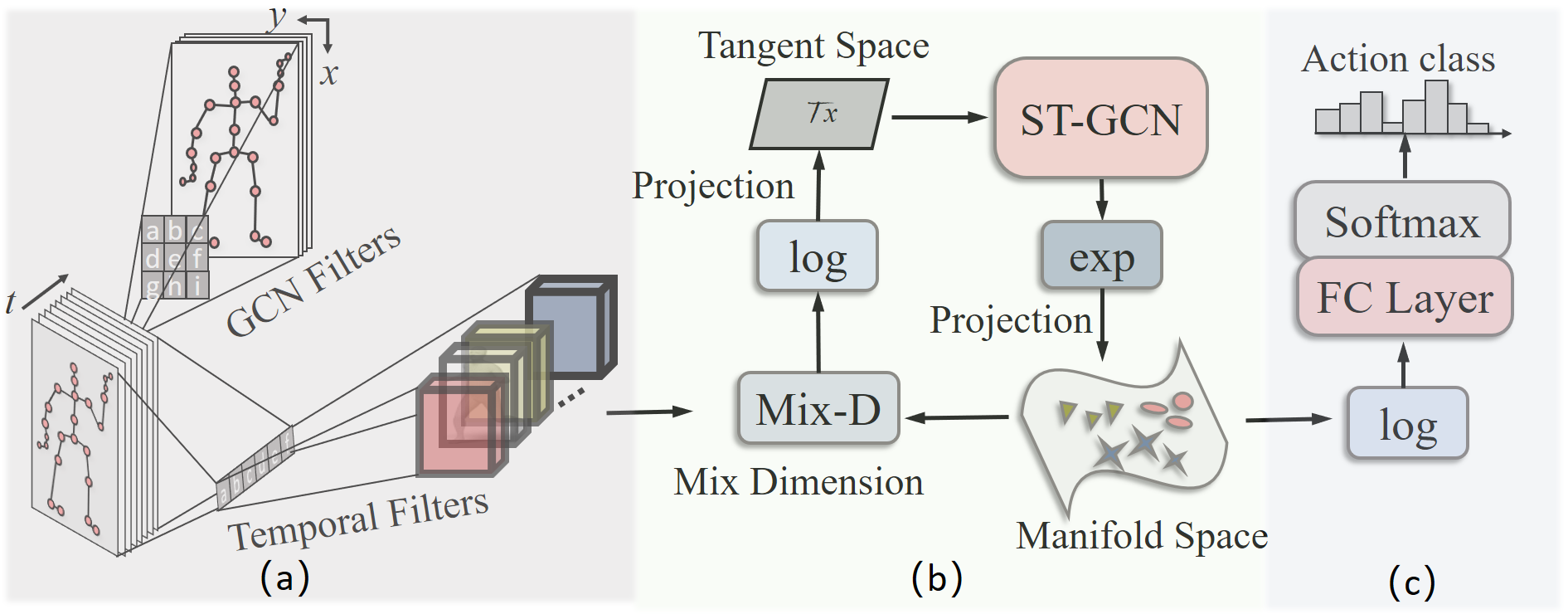}
    \caption{Illustration of our framework. There are mainly three stages in our framework,  including (a) Graph feature embedding, (b) Extracting graph in Poincar\'e model, and (c) Classification in the Euclidean space. At the first stage, we utilize GCN filters to capture the graph representation for each frame and temporal filters are then used to capture the dynamic information. With this output from the first stage, we mix different dimensions and then project them to the tangent space, where an ST-GCN is used to extract higher-level graph representations. The feature then is mapped back to the manifold. In this way, we model the graph on the Riemannian manifold. Note here, the ST-GCN is based on the same module in stage (a) and the manifold space here is on Poincar\'e model. We stack several modules from stage (b) to capture higher-level semantic representations. After that, as shown in stage (c), the graph feature is projected back to Euclidean space such that a Euclidean loss function can be used to optimize this process.}
    \label{fig:my_framework}
\end{figure*}

In this section, we describe our ST-GCN defined on Poincar\'e model. The framework is illustrated in Figure~\ref{fig:my_framework}. The basic module of our network consists of a GCN followed by a temporal convolutional filter. Such block is stacked for multiple layers to capture the high-level representations for the graph sequences. The spatial temporal model is defined by the Poincar\'e geometry. Finally, the learned features are projected back to Euclidean space for the prediction. We will detail the important components of our framework in the following parts.

\subsection{Spatial-Temporal Graph Convolutional Networks}

Graph data is a very useful data structure to model complex structures such as social network, and bio-information, while it is not easy to model this kind of irregular data using neural networks. Define $\mathcal{G}$ as one frame of the skeleton sequence. Assume the skeleton is composed of $N$ nodes and the node connections are encoded in the adjacency matrix $A \in \mathit{R}^{N \times N}$. $X \in \mathit{R}^{N\times 3}$ denotes the input representations for the graph (skeleton) with $N$ nodes. To extract the feature representation of $\mathcal{G}$, a Fourier transform is conducted on it such that the transformed signal could then be processed with formulation of fundamental operations such as filtering. As a result, a normalized graph Laplacian $L = I_n - D^{-1/2}AD^{-1/2}$ is used for Fourier transform. Here, the diagonal degree matrix $D$ is constructed with elements $D_{ii} = \sum_{j}A_{ij}$. Then a graph filtered by operator $g_{\theta}$, parameterized by $\theta$, can be formulated as
\begin{equation}
    Y = g_{\theta}(L)X = U g_{\theta}(\Lambda) U^{T}X,
\end{equation}

\noindent where $Y$ is the extracted graph feature and $U$ is the Fourier basis subjected to $L = U \Lambda U^T$ with the $\Lambda$ as its corresponding eigenvalues. Suggested by~\cite{hammond2011wavelets}, the filter $g_{\theta}$ can be further approximated by a Chebyshev polynomials with $K$-th order such that the computational burden is significantly reduced. That is
\begin{equation}\label{eq:cheb_approximation}
    Y = \sum_{k=0}^{K}\theta_k^{'}T_k(\hat{L})X,
\end{equation}

\noindent where $\theta_k^{'}$ denotes Chebyshev coefficients. For $k>1$, Chebyshev polynomial $T_k(\hat{L})$ is recursively defined as
\begin{equation}\label{eq:cheb}
    T_k(\hat{L}) = 2\hat{L}T_{k-1}(\hat{L})-T_{k-2}(\hat{L})
\end{equation}
with $T_0 = 1$ and $T_1 = \hat{L}$. Here $\hat{L} = 2L/{\lambda_{max}}-I_n$ is normalized to [-1,1]. For Eq.~(\ref{eq:cheb_approximation}), work in~\cite{kipf2016semi} sets $K=1$, $ \lambda_{max}= 2$. In this way, a first-order approximation of spectral graph convolutions is formed. Therefore, 
\begin{equation}
    Y = \theta_0^{'}X +\theta_1^{'}(L-I_n)X = \theta_0^{'}X -\theta_1^{'}(D^{-1/2}AD^{-1/2})X.
\end{equation}

\noindent This equation can be further simplified by using an unified parameter $\theta$, which means setting $\theta = \theta_0 = - \theta_1$ and making the training process adapt the approximation error, then 
\begin{equation}
    Y = \theta(I_n + D^{-1/2}AD^{-1/2})X.
\end{equation}
We set  $ L = I_n + D^{-1/2}AD^{-1/2}$ in the following parts for simplicity. An ST-GCN is then built by succeeding a temporal filter after each GCN. One can stack multiple GCN layers to get high-level graph features. Generally, at middle GCN layers, $X\in \mathcal{R}^{n\times C}$ is with multi-channels. Thus, we get
\begin{equation}\label{eq:gcn}
    Y = LX\theta.
\end{equation}

To further improve the robustness, Works in~\cite{shi2019two,peng2020learning} present an ST-GCN block which generates dynamic embedding matrix based on the node correlations.  Instead of providing a dynamic one, we design our ST-GCN block in Poincar\'e model, making the node representations fit with original graph structure. In this way, we capture the graph features with much less parameters.

\subsection{Poincar\'e Manifold}

\begin{figure}
    \centering
    \includegraphics[width=0.35\textwidth]{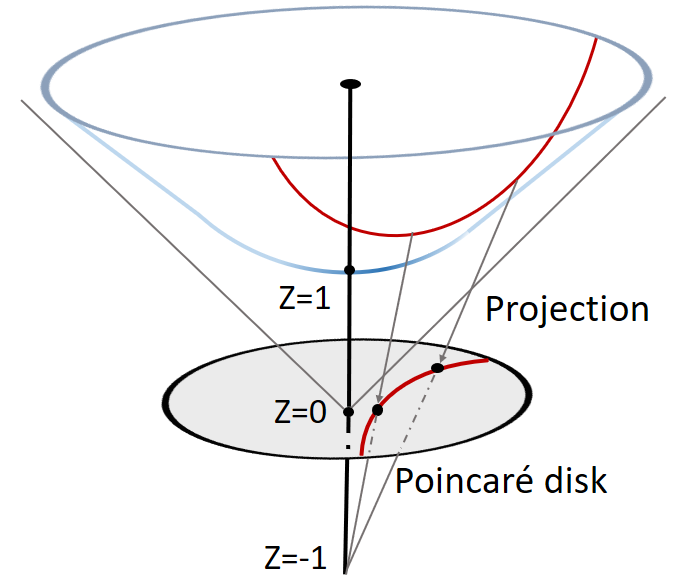}
    \caption{Illustration of the 2D Poincar\'e model. For any point on the hyperboloid, we form a line by extending it out to a focal point (0,0,-1). Then, the intersection point on the z=0 plane will be its projection point in the Poincar\'e model.}
    \label{fig:poincare}
\end{figure}

\begin{figure}
    \centering
    \includegraphics[width=0.25\textwidth]{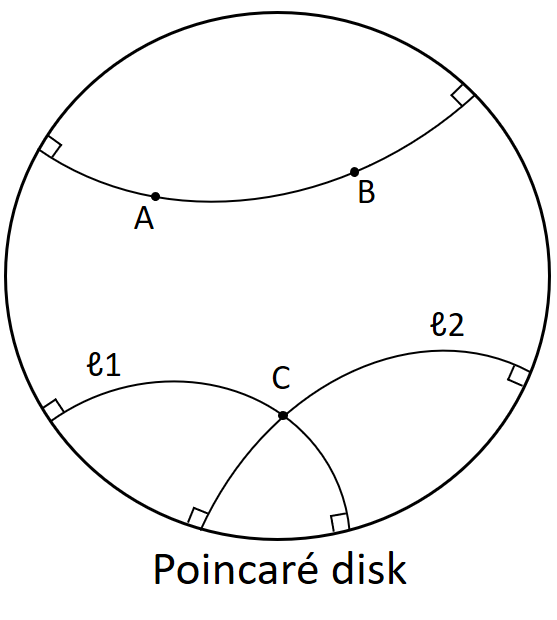}
    \caption{Illustration of the property of Poincar\'e model. Go through point C, one can draw more than one line ($l_1$ and $l_2$) which is paralleled to line $AB$.}
    \label{fig:poincare2}
    \vspace{-2em}
\end{figure}

In this section, we will then discuss how to define an ST-GCN on a Riemannian manifold. Here, we focus on the Poincar\'e model in the Hyperbolic space~\cite{reynolds1993hyperbolic}, which is a maximally symmetric, simply connected Riemannian manifold with constant negative sectional curvature. Hyperbolic space is analogous to the $n$-dimensional sphere, which has constant positive curvature.  As a special case, the Poincar\'e model can be derived using a stereoscopic projection of the hyperboloid model onto the unit circle of the z=0 plane. It is hard to visualize as it is hard to imagine in a curved space. We can embed models of 2D hyperbolic geometry into a pseudo-Euclidean space called Minkowski space~\cite{tataru2001strichartz}. Here, a $n$-dimensional Minkowski space is a real vector space of real dimension n, in which there is a constant Minkowski metric. As illustrated in the Figure~\ref{fig:poincare}, the Poincar\'e disk i.e., a 2D Poincar\'e model, is constructed as a projection from the upper half hyperboloid onto the unit disk at z=0. Poincar\'e disk breaks the rule in Euclidean space. For instance, like in the Figure~\ref{fig:poincare2}, given a line $\overleftrightarrow{AB}$ and a point $C \notin \overleftrightarrow{AB}$, then we can draw at least two lines cross through $C$ that do not intersect line $\overleftrightarrow{AB}$.  The two lines through $C$, denoted as lines $\mathit{l}1$ and $\mathit{l}2$. Different from the Euclidean 2D space, we have that $\overleftrightarrow{AB}$ parallels both to $\mathit{l}1$ and $\mathit{l}2$, but at the same time $\mathit{l}1$ and $\mathit{l}2$ are not paralleled. Note also that what different from the Euclidean space is that $\mathit{l}2$ intersects one of a pair of parallel lines ($\mathit{l}1$), but does not intersect the other parallel line ($\overleftrightarrow{AB}$). In hyperbolic geometry, a significant intrinsic property is the exponential growth, instead of the polynomial growth as in Euclidean geometry. This means that the distance between irrelevant samples will be exponentially greater than the distance between similar samples. Thus, the relationship between samples represented in the hyperbolic space can emphasize similar samples and suppress irrelevant samples. As a result, hyperbolic geometry outperforms Euclidean geometry in some special tasks, such as learning hierarchical embedding~\cite{ganea2018hyperbolic}.

Here, we formally define this manifold. Let $\mathcal{M}$ be a $n$-dimensional manifold. There are three basic components which count much important for the manifold $\mathcal{M}$, namely, geodesic, tangent space, and the Riemannian metric. 

A geodesic is the generalization of a straight line to curved space, defined to be a curve where you can parallel transport a tangent vector without deformation. In our hyperboloid model, as illustrated in Fig.~\ref{fig:poincare}, the geodesic (or our hyperbolic line) is defined to be the curve created by intersecting the plane defined by two points and the origin (i.e., coordinate (0,0,0)) with the hyperboloid. So one point end has to go down first then back up to reach another point. This distance is not directly towards it using the shortest path on the surface in Euclidean space, but will be around the circumference. Formally, the distance for $x,y \in \mathcal{M}$ are defined as:
\begin{equation}
    d(x,y) = arcosh(1+2\frac{||x-y||^2}{(1-||x||^2)(1-||y||^2)}).
\end{equation}

A tangent space $\mathcal{T}_{x}\mathcal{M}$ at point $x$ is defined as the first order linear approximation of $\mathcal{M}$ around $x$.
A Riemannian metric $g$ is a collection of inner-products $g_x: \mathcal{T}_{x}\times \mathcal{T}_{x}\rightarrow \mathit{R}$ varying smoothly with $x$. For the Poincar\'e model, the open unit ball equipped with the Riemannian metric tensor, which is defined with
\begin{equation}
    g_x = (\frac{2}{1-||x||^{2}})^{2}g^{\mathit{E}}
\end{equation}
where $g^{E} = \mathbf{I}_n$ denotes the Euclidean metric tensor, which is conformal to the Euclidean one. Then, a Riemannian manifold $(\mathcal{M},g)$ is a manifold $\mathcal{M}$ with a group Riemannian metric $g$. 

Now, we build our spatial-temporal graph convolutional networks on the Poincar\'e geometry to provide a more robustness representation for the temporal graph sequences. One advantage of the hyperbolic space is that it provides a bijection between the hyperbolic space and the tangent space at a point such that the operations for points on the hyperboloid manifold can be performed in tangent space and then mapped back and vice-versa. The bijection is done by the exponential map, which maps the points on the tangent space to the manifold and is defined as $\text{exp}_x: \mathcal{T}_{x}\mathcal{M}\rightarrow \mathcal{M}$. The logarithmic map, which, as the inverse step, maps points on the tangent space back to the manifold, is defined as $\text{log}_x: \mathcal{M} \rightarrow  \mathcal{T}_{x}\mathcal{M}$. Mathematically,
\begin{equation}
    \text{exp}_x(v) = x \oplus (tanh(\frac{\lambda_x||v||}{2})\frac{v}{||v||})
\end{equation}
    
\begin{equation}
    \text{log}_x(y) = \frac{2}{\lambda_x}archtanh(||-x\oplus y||)\frac{-x\oplus y}{||-x\oplus y||}
\end{equation}

\noindent where $v$ is a tangent vector, and $\lambda_x = \frac{2}{1-||x||^{2}}$ is a conformal factor. $\oplus$ is the Möbius addition for any $x,y \in \mathcal{M}$:
\begin{equation}
    x\oplus y = \frac{(1+2\left \langle x,y  \right \rangle+||y||^2)x+(1-||x||^2)y}{1+2\left \langle x,y  \right \rangle +||x||^2||y||^2}.
\end{equation}

With the aforementioned projection functions, we perform the GCN operations on the Poincar\'e models. Since there is no definition of vector space in the Riemannian space, inspired by~\cite{ganea2018hyperbolic}, we conduct the feature extraction on the tangent space by projecting the graph embeddings with the logarithmic map. In this way, the Euclidean transformation conducted by the neural operations can be utilized for the feature in hyperbolic space. Specifically, as shown in Fig.~\ref{fig:my_framework}, there are three stages in our framework. The input raw data is first encoded by a feature embedding network. Then, the captured feature is projected into a tangent space and filtered by a GCN and a temporal filter, which are
\begin{equation}
    Y_i = \mathit{L}~\text{log}_x(\mathit{X_i})\Theta,
\end{equation}
\begin{equation}
    Y = \Theta_{t\times 1}\{Y_i\}^T,
\end{equation}
Here, $X_i$ represents the $i$-th frame of the inputs and $\{Y_i\}^T$ are the representations of the entire T frames after the GCN. $\{Y_i\}^T$ are then fed into a temporal filter with kernel $t \times 1$. There will also be an activation function to perform a non-linear projection on $Y$. Finally, we map the representation back to the a Euclidean space with $\text{log}_x$ function and thus optimize the network with Euclidean loss function. 

\subsection{Mix-Dimensions on Manifold}

With aforementioned method, we can build an ST-GCN on Riemannian manifold. However, it is still not easy to manually determine the projection dimension for each layer on the manifold. There will be thousands of combinations for a ten layers ST-GCN, which definitely is not possible to manually evaluate each setting and find out the best one. Here, we provide an efficient way to explore the optimal settings for each layer. Inspired by the slimmable networks~\cite{yu2018slimmable}, where they trained a slimmable network to perform the network pruning, we mix different dimensions at each projection point to efficiently explore the best projection dimension for each layer. Specifically, we provide a set of projection dimensions on the Poincar\'e model. Thus, there will be a group of corresponding ST-GCN blocks on the tangent space. Instead of constructing a set of ST-GCN blocks there, we make the higher dimension projection share the operations of the lower ones. In this way, we can build a super-model and the exploration for the higher dimension would not need training from scratch since they could benefit from the training for the lower dimension. To this end, we construct corresponding switchable batch normalization~\cite{yu2018slimmable} and a slimmable network on the tangent space.

In fact, this mix-dimension method provides thousands of combinations of the ST-GCNs on the Poincar\'e model at one time. Instead of evaluating the performance of all these combinations, we compute the relative modeling ability for this task. We assume that the super-model based on the mix-dimension method could provide an estimation of each individual dimension setting. As a result, we build the ST-GCN model by dividing each layer into specific group, and we try to figure out what is the best combination for this network. For instance, at a layer, we project the graph representation into Poincar\'e model with 64 dimensions. Instead of only providing this projection,we project the graph to a set of dimensions,$[32, 48, 64, 80, 96]$, at the same time. We conduct the same processing for the other layers. Therefore, there will be five different ST-GCN blocks for this single layer. Thus, However, it will cost expensive computation for a deep GCN models. So instead of building all the models, we construct network with a slimmbale biggest one. In each iteration, we randomly sample a combination of projection dimension and the corresponding ST-GCN will be activated and trained. After the training phase is finished, we randomly choose a batch of projection combinations. Based on their relative prediction accuracy on this task, we choose the best one as our ST-GCN on the Poincar\'e model. 

\section{Experiments}\label{sec:exp}
This section describes the experiments in terms of datasets, the architecture, the training details, the comparison results and the corresponding analysis.

\begin{figure}
    \centering
    \includegraphics[width=0.4\textwidth]{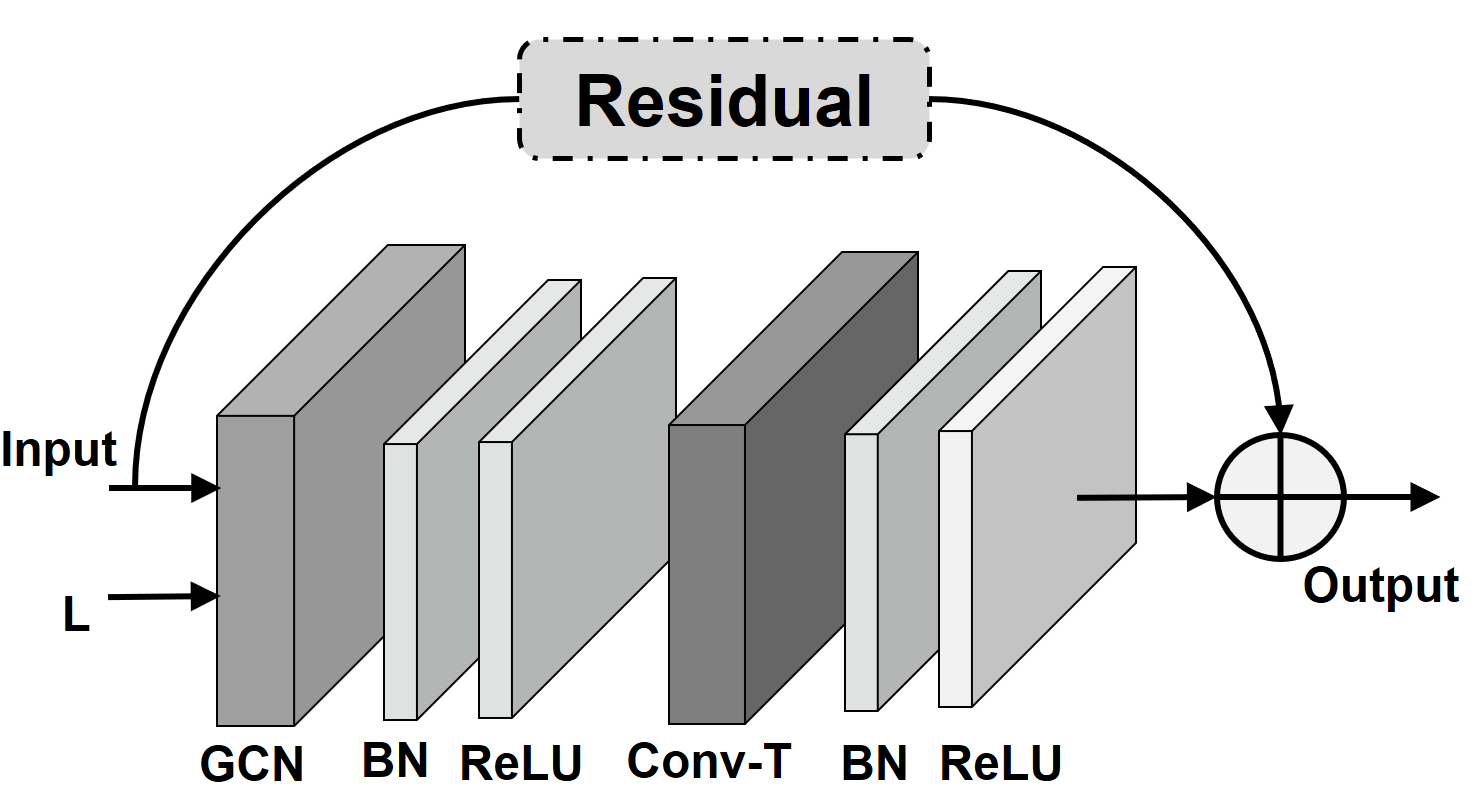}
    \caption{ Illustration of ST-GCN block. There are two inputs, including the $\textbf{L}$ and the $\textbf{X}$ in Eq.~(\ref{eq:gcn}). Here, GCN denotes the spatial graph convolutional network, and Conv-T represents the temporal filter. Both of them are followed by batch normalization (BN) layer and a activation layer (ReLU). Moreover, a residual connection is added for each block. The output and the original $\textbf{L}$ are fed into next block.}
    \label{fig:stgcn}
\end{figure}
\subsection{Datasets}
In the experiments, we evaluate our method on two current largest scale 3D skeleton-based action recognition datasets, i.e., NTU RGB+D~\cite{shahroudy2016ntu} and NTU RGB+D 120~\cite{liu2019ntu}.

\textbf{NTU RGB+D}~~NTU RGB+D~\cite{shahroudy2016ntu} is currently one of the most widely used action recognition datasets. There are four modalities, including RGB videos, depth sequences, infrared videos and 3D skeleton data. Here, we only conduct experiments on the skeleton data. There are totally 56,880 video clips involving 60 human action classes and the max number of frames in each sample is 300. They are captured from three cameras with different settings. In this dataset, each actor skeleton contains 25 3D joints coordinates and there are at most two actors in each video clip. The original benchmark provided two evaluations, which are Cross-Subject (X-subject) and the Cross-View (X-view) evaluations. In the X-subject evaluation, the training set contains 40,320 videos from 20 subjects, and the rest 16,560 video clips are used for testing. In the X-view evaluation, 37,920 videos captured from cameras No.2 and No.3 are used in the training and the rest 18,960 videos from camera No.1 are used for testing. We follow the original two benchmarks and report the Top-1 accuracy. For inputs with more than one stream, e.g., bones, a score-level fusion result is reported.

\textbf{NTU RGB+D 120}~~~~NTU RGB+D 120~\cite{liu2019ntu} is the extended version of the aforementioned NTU RGB+D dataset. It is a more challenging dataset with high variability since it involves more subjects and action categories. There are totally 114,480 samples which cover 120 classes. They are captured from 106 distinct subjects in a wide range of age distribution. There are two different evaluation protocols in this new released dataset: Cross-Subject (X-subject), like in the previous dataset, which splits subjects in half to training and testing parts; Cross-Setup (X-setup), which divides samples by 32 camera setup IDs. The even setup IDs is utilized for training (16 setups) and odd setup IDs for testing (16 setups). Here, 32 IDs are based on different camera setups, e.g., different heights with different horizontal angles. In consistent with the original benchmark method, Top-1 accuracy is reported in the two benchmarks. 

Before the data was fed into the networks, there should be some pre-processing such that the data structure for each video clip is unified. We keep it consistent with the previous best methods~\cite{shi2019two,peng2020learning} for fair comparison. All the skeleton sequences are unified to 300 frames with two actors. For the single-actor data, the second body  will be padded with all zeros. Each dimension of the 3D coordinates is put into three channels as inputs. Like~\cite{shi2019two,peng2020learning}, we also compute the second-order information of joint, i.e., bones.

\subsection{Architecture and Training}

The basic framework is as illustrated in Figure~\ref{fig:my_framework}. Firstly, we build an embedding model with the original ST-GCN module (like in Fig.~\ref{fig:stgcn}), in which the graph convolutional layer is followed by a temporal convolution with a kernel size $9\times 1$ to capture the dynamic information. The three dimensional node coordinates are projected into a space with 64D.   Then, as shown in the second stage of the framework, we build networks on manifold. A group of manifold projection, [32, 48, 64, 80, 96], is constructed. The corresponding ST-GCN layers are also built on Poincar\'e model. We empirically stack six layers on this manifold space and divide them into three levels. For each level, we insert two ST-GCN layers and double the projection group setting based on the previous level, which means the mixed dimensions at second and third level are [64, 96, 128, 160, 192], and [128, 192, 256, 320, 384], respectively. Finally, as shown in the third stages of Fig.~\ref{fig:my_framework}, the graph representations modeled from the manifold space is projected back to the Euclidean space. The resulted node embeddings are averaged to a feature vector. Then a fully connected layer followed by a softmax function is utilized to predict a class prediction. 

There are mainly two steps to perform the training. Firstly, we mix dimensions and efficiently explore the best combination on the NTU RGB+D data, X-subject evaluation, for 50 epochs. Then we randomly sample 100 combinations and choose the best one based on the prediction accuracy. We find that one of the best projection combination for the six layers is [64, 64, 160, 128, 320, 256]. Then, based on this resulted combination, we remove the mix-dimension process and build our ST-GCN on Poincar\'e model. During the training process, the cross-entropy loss is utilized as the classification loss. The learning rate is set as 0.1 and is decreased based on a cosine function. A stochastic gradient descent (SGD) with Nesterov momentum (0.9) is applied as the optimization algorithm for the network. We set the weight decay to 0.0005 as regularization. This model will be trained for 50 epochs and compared to other approaches.

\subsection{Ablation Study}
In this section, we evaluate the effectiveness of our method on the NTU RGB+D dataset under the given two evaluation metrics. Here, current state-of-the-art ST-GCN~\cite{yan2018stgan} works as a baseline since we share the same ST-GCN module. To further evaluate that how much we can gain from the Poincar\'e geometry,  we implement a smaller ST-GCN (seven layers with output channels [64, 64, 64, 128, 128, 256, 256] for each one) by keeping its network architecture totally consistent with Ours. This model is referred as ST-GCN-s. Finally, we employ the optimal projection dimensions on the manifold explored by our mix-dimension method. This network is represented by Ours-M.

The comparison results are listed in Table~\ref{tab:ablation}. It can be seen from this table that for the given two evaluations, our networks can largely improve the performance on both joint and bone data. And with the helps of our Mix-dimension, the performance of the proposed model can be further improved. Specifically, under the X-subject evaluations, our model (Ours-M) can overcome the baseline from joint and bone data by 7.6\% and 9.6\%, respectively. When compared to ST-GCN-s, which is with a similar model size to Ours, the proposed model defined with Poincar\'e model could even outperform it by 8.6\% and 10.4\%, respectively. All of them proves that defining ST-GCN on manifold space could benefit greatly. 

\begin{table}[t]\footnotesize
\begin{center}
\caption{\small{\textbf{Ablation Study}. Performance comparison on NTU RGB+D dataset under two given evaluations. Here we compare with ST-GCN~\cite{yan2018stgan} and a small ST-GCN (seven layers), denoted as ST-GCN-s, by employing same architecture with Ours. Here, Ours-M is the architecture resulted from Mix-Dimension. The big improvements achieved by our method prove the effectiveness of our model.}}
\vspace{3mm}
\label{tab:ablation}
\scalebox{0.85}{
\begin{tabular}{l| l |c |c }

\toprule

&Methods &~Joint ~&~Bone ~ \\
\hline 
\multirow{3}{*}{\textbf{X-subject~~}}
& ST-GCN~\cite{yan2018stgan}  &80.2\%&77.6\%\\
&ST-GCN-s  &79.2\%&76.8\%\\
&\textbf{Ours}  & \textbf{86.2\%}&\textbf{85.7\%}\\
&\textbf{Ours-M}  & \textbf{87.8\%}&\textbf{87.2\%}\\
\midrule
\midrule
\multirow{3}{*}{\textbf{X-view}}
&ST-GCN~\cite{yan2018stgan}  &91.1\%&90.5\%\\
&ST-GCN-s  &89.7\%&88.3\%\\

&\textbf{Ours}  &\textbf{93.9\%}&\textbf{93.7\%}\\
&\textbf{Ours-M}  &\textbf{95.0\%}&\textbf{94.7\%}\\
\bottomrule
\end{tabular}}
\end{center}
\vspace{-2em}

\end{table}

\subsection{Comparison with the State-Of-The-Art methods} 

First, we conduct experiments on NTU RGB+D dataset, and like~\cite{shi2019two,peng2020learning}, we report the best results after performing the score-level fusion on joints and bones. Here, we compared with 14 State-Of-The-Art (SOTA) approaches under the given evaluation metrics. There is one hand-crafted method~\cite{hu2015jointly}, but most of them are deep learning methods, including CNN-based method~\cite{kim2017interpretable}, RNN-based methods~\cite{shahroudy2016ntu,song2017end,zhang2017view,si2018skeleton} and reinforcement learning based method~\cite{tang2018deep}. Besides, we also compare with various ST-GCNs~\cite{li2018spatio,yan2018stgan,li2019spatio,gao2019optimized,li2019actional,shi2019two,peng2020learning}, which get very promising results on this task. It is also worth mentioning that Peng \etal~\cite{peng2020learning} introduced a neural architecture search (NAS) method into this task and automatically design an ST-GCN, which obtains the current best results.  We also compare with this method. All the comparison results are listed in Table~\ref{tab:NTU}. It can be seen from Table~\ref{tab:NTU} that under both X-subject and X-view evaluations, our model achieves the current best classification accuracy. Specifically, compared with current best method NAS-GCN~\cite{peng2020learning}, our model can achieve better results and also decrease the model size by 2.5 times. 

\begin{table}[t]\footnotesize
\begin{center}
\caption{ \small{Performance comparison on NTU RGB+D with 14 state-of-the-art methods. }}
\vspace{3mm}
\label{tab:NTU}
\scalebox{0.9}{
\begin{tabular}{lc c}

\toprule
Methods &X-subject &X-view \\
\hline 
Dynamic Skeleton~\cite{hu2015jointly} &60.2\% &65.2\%\\
Part-Aware LSTM~\cite{shahroudy2016ntu} &62.9\% &70.3\% \\
STA-LSTM~\cite{song2017end} &73.4\% &81.2\% \\
TCN~\cite{kim2017interpretable} &74.3\% &83.1\% \\
VA-LSTM~\cite{zhang2017view} &79.2\% &87.7\% \\
Deep STGCK~\cite{li2018spatio}& 74.9\%& 86.3\% \\
ST-GCN~\cite{yan2018stgan} &81.5\% &88.3\% \\
DPRL~\cite{tang2018deep} &83.5\% &89.8\% \\
SR-TSL~\cite{si2018skeleton} & 84.8\% &92.4\% \\
STGR-GCN~\cite{li2019spatio}& 86.9\% &92.3\% \\
GR-GCN~\cite{gao2019optimized}  &87.5\% &94.3\%\\
AS-GCN~\cite{li2019actional}  &86.8\% &94.2\%\\
2S-AGCN~\cite{shi2019two}  &88.5\% &95.1\%\\
NAS-GCN~\cite{peng2020learning}  &89.4\% &95.7\%\\
\midrule

\textbf{Ours-M}  &\textbf{89.7\%}& \textbf{96.0\%}\\
\bottomrule
\end{tabular}}
\end{center}

\end{table}

On NTU RGB+D 120 dataset, we compared to 14 skeleton-based action recognition approaches under X-subject and X-setup evaluation metrics. The methods includes hand-crafted method~\cite{hu2015jointly}, conventional deep learning methods, including Part-Aware LSTM~\cite{shahroudy2016ntu}, Soft RNN~\cite{hu2018early}, Spatio-Temporal LSTM~\cite{liu2016spatio}, Internal Feature Fusion~\cite{liu2017skeleton}, GCA-LSTM~\cite{liu2017global}, Multi-Task Learning Network~\cite{ke2017new}, FSNet~\cite{liu2019skeleton} , Skeleton Visualization~\cite{liu2017enhanced}, Two-Stream Attention LSTM~\cite{liu2017skeleton2}, Multi-Task CNN with RotClips~\cite{ke2018learning}, and Body Pose Evolution Map~\cite{liu2018recognizing}. We also compare with the promising GCN-based models~\cite{yan2018stgan,li2019actional}. This task is more challenging than the above one, thus all the baselines are lower than 80\%. Here, we report the best result on joint data. All the comparison results are listed in Table~\ref{tab:NTU-120}.

We can see from Table~\ref{tab:NTU-120} that graph convolutional networks are much suitable for this task when comparing the last two methods to previous hand-crafted method and conventional deep learning methods. Under the X-subject protocol, our model improves previous best CNN-based method~\cite{liu2018recognizing} by more than 15\%. Compared to the GCN methods, we can still get 4.3\% and 2.8\% improvements under the X-setup and X-subject evaluation metrics, respectively. 
\begin{table}[thbp!]\footnotesize
\begin{center}
\caption{ \small{Performance comparison on NTU RGB+D 120 with 13 state-of-the-art methods. $*$ means implemented by us.}}
\vspace{0.5em}
\label{tab:NTU-120}
\scalebox{0.85}{
\begin{tabular}{l c c }
\toprule
Methods &X-setup  &X-subject \\
\hline 
Part-Aware LSTM~\cite{shahroudy2016ntu} &25.5\% & 26.3\% \\
Soft RNN~\cite{hu2018early} & 36.3\% & 44.9\% \\
Dynamic Skeleton~\cite{hu2015jointly} & 50.8\% & 54.7\% \\
Spatio-Temporal LSTM~\cite{liu2016spatio} & 55.7\% & 57.9\% \\
Internal Feature Fusion~\cite{liu2017skeleton} & 58.2\% & 60.9\% \\
GCA-LSTM~\cite{liu2017global} & 58.3\% & 59.2\% \\
Multi-Task Learning Network~\cite{ke2017new} & 58.4\% & 57.9\% \\
FSNet~\cite{liu2019skeleton} &59.9\% &62.4\% \\
Skeleton Visualization~\cite{liu2017enhanced} &60.3\% &63.2\% \\
Two-Stream Attention LSTM~\cite{liu2017skeleton2} &61.2\% &63.3\% \\
Multi-Task CNN with RotClips~\cite{ke2018learning} &62.2\% &61.8\% \\
Body Pose Evolution Map~\cite{liu2018recognizing} &64.6\% & 66.9\% \\
ST-GCN~\cite{yan2018stgan} &71.3\% & 72.4\% \\
AS-GCN~\cite{li2019actional} &78.9\% & 77.7\% \\
\midrule
\textbf{Ours-M}  &\textbf{83.2\%}& \textbf{80.5\%}\\
\bottomrule
\end{tabular}}
\end{center}
\vspace{-2em}

\end{table}

\section{Conclusions}\label{sec:conclusion}

In this paper, we address the task of video action recognition from skeleton sequences. To better modelling the graph-structure data, we propose a novel spatial-temporal graph convolutional network by applying Poincar\'e geometry. The graph embedding is projected into a Poincar\'e model where we further learn and extract a high-level of graph representation. To further explore the optimal projection dimension on the manifold space and provide a better setting for different layers, we provide an efficient exploration strategy by mixing a group of dimensions on the Poincar\'e model. With the resulted dimension setting for each layer, we construct our ST-GCN on this manifold space. Then, we conduct extensive comparison experiments to verify the effectiveness of our method. Comparison results on two current most challenging datasets show that the features modelled by our model are much more distinguishable than the model with the same setting in Euclidean space. Besides, our model can also get the best results for the given tasks with only 40\% of the previous best model in term of the model size. All of these prove the effectiveness of our model.

\section{Acknowledgements}
This work was supported by ICT 2023 project (grant 328115), the Academy of Finland for project MiGA (grant 316765), and Infotech Oulu. As well, the authors wish to acknowledge
CSC-IT Center for Science, Finland, for computational resources.

\bibliographystyle{aaai}
\bibliography{aaai19-main}

\begin{thebibliography}{}

\bibitem[\protect\citeauthoryear{Benedetti and
  Petronio}{2012}]{benedetti2012lectures}
Benedetti, R., and Petronio, C.
\newblock 2012.
\newblock {\em Lectures on hyperbolic geometry}.
\newblock Springer Science \& Business Media.

\bibitem[\protect\citeauthoryear{Bronstein \bgroup et al\mbox.\egroup
  }{2017}]{bronstein2017geometric}
Bronstein, M.~M.; Bruna, J.; LeCun, Y.; Szlam, A.; and Vandergheynst, P.
\newblock 2017.
\newblock Geometric deep learning: going beyond euclidean data.
\newblock {\em IEEE Signal Processing Magazine} 34(4):18--42.

\bibitem[\protect\citeauthoryear{Chami \bgroup et al\mbox.\egroup
  }{2019}]{chami2019hyperbolic}
Chami, I.; Ying, Z.; R{\'e}, C.; and Leskovec, J.
\newblock 2019.
\newblock Hyperbolic graph convolutional neural networks.
\newblock In {\em Advances in Neural Information Processing Systems},
  4869--4880.

\bibitem[\protect\citeauthoryear{Cho and Lee}{2017}]{cho2017riemannian}
Cho, M., and Lee, J.
\newblock 2017.
\newblock Riemannian approach to batch normalization.
\newblock In {\em Advances in Neural Information Processing Systems},
  5225--5235.

\bibitem[\protect\citeauthoryear{Defferrard, Bresson, and
  Vandergheynst}{2016}]{defferrard2016convolutional}
Defferrard, M.; Bresson, X.; and Vandergheynst, P.
\newblock 2016.
\newblock Convolutional neural networks on graphs with fast localized spectral
  filtering.
\newblock In {\em Advances in neural information processing systems},
  3844--3852.

\bibitem[\protect\citeauthoryear{Du, Wang, and Wang}{2015}]{du2015hierarchical}
Du, Y.; Wang, W.; and Wang, L.
\newblock 2015.
\newblock Hierarchical recurrent neural network for skeleton based action
  recognition.
\newblock In {\em Proceedings of the IEEE conference on computer vision and
  pattern recognition},  1110--1118.

\bibitem[\protect\citeauthoryear{Ganea, B{\'e}cigneul, and
  Hofmann}{2018}]{ganea2018hyperbolic}
Ganea, O.; B{\'e}cigneul, G.; and Hofmann, T.
\newblock 2018.
\newblock Hyperbolic neural networks.
\newblock In {\em Advances in neural information processing systems},
  5345--5355.

\bibitem[\protect\citeauthoryear{Gao \bgroup et al\mbox.\egroup
  }{2019}]{gao2019optimized}
Gao, X.; Hu, W.; Tang, J.; Liu, J.; and Guo, Z.
\newblock 2019.
\newblock Optimized skeleton-based action recognition via sparsified graph
  regression.
\newblock In {\em Proceedings of the 2019 ACM International Conference on
  Multimedia}.
\newblock ACM.

\bibitem[\protect\citeauthoryear{Gulcehre \bgroup et al\mbox.\egroup
  }{2018}]{gulcehre2018hyperbolic}
Gulcehre, C.; Denil, M.; Malinowski, M.; Razavi, A.; Pascanu, R.; Hermann,
  K.~M.; Battaglia, P.; Bapst, V.; Raposo, D.; Santoro, A.; et~al.
\newblock 2018.
\newblock Hyperbolic attention networks.
\newblock {\em arXiv preprint arXiv:1805.09786}.

\bibitem[\protect\citeauthoryear{Hammond, Vandergheynst, and
  Gribonval}{2011}]{hammond2011wavelets}
Hammond, D.~K.; Vandergheynst, P.; and Gribonval, R.
\newblock 2011.
\newblock Wavelets on graphs via spectral graph theory.
\newblock {\em Applied and Computational Harmonic Analysis} 30(2):129--150.

\bibitem[\protect\citeauthoryear{Hu \bgroup et al\mbox.\egroup
  }{2015}]{hu2015jointly}
Hu, J.-F.; Zheng, W.-S.; Lai, J.; and Zhang, J.
\newblock 2015.
\newblock Jointly learning heterogeneous features for rgb-d activity
  recognition.
\newblock In {\em IEEE Conference on Computer Vision and Pattern Recognition},
  5344--5352.

\bibitem[\protect\citeauthoryear{Hu \bgroup et al\mbox.\egroup
  }{2018}]{hu2018early}
Hu, J.-F.; Zheng, W.-S.; Ma, L.; Wang, G.; Lai, J.-H.; and Zhang, J.
\newblock 2018.
\newblock Early action prediction by soft regression.
\newblock {\em IEEE transactions on pattern analysis and machine intelligence}.

\bibitem[\protect\citeauthoryear{Ke \bgroup et al\mbox.\egroup
  }{2017}]{ke2017new}
Ke, Q.; Bennamoun, M.; An, S.; Sohel, F.; and Boussaid, F.
\newblock 2017.
\newblock A new representation of skeleton sequences for 3d action recognition.
\newblock In {\em Proceedings of the IEEE conference on computer vision and
  pattern recognition},  3288--3297.

\bibitem[\protect\citeauthoryear{Ke \bgroup et al\mbox.\egroup
  }{2018}]{ke2018learning}
Ke, Q.; Bennamoun, M.; An, S.; Sohel, F.; and Boussaid, F.
\newblock 2018.
\newblock Learning clip representations for skeleton-based 3d action
  recognition.
\newblock {\em IEEE Transactions on Image Processing} 27(6):2842--2855.

\bibitem[\protect\citeauthoryear{Kim and Reiter}{2017}]{kim2017interpretable}
Kim, T.~S., and Reiter, A.
\newblock 2017.
\newblock Interpretable 3d human action analysis with temporal convolutional
  networks.
\newblock In {\em 2017 IEEE Conference on Computer Vision and Pattern
  Recognition workshops},  1623--1631.

\bibitem[\protect\citeauthoryear{Kipf and Welling}{2016}]{kipf2016semi}
Kipf, T.~N., and Welling, M.
\newblock 2016.
\newblock Semi-supervised classification with graph convolutional networks.
\newblock {\em arXiv}.

\bibitem[\protect\citeauthoryear{Li \bgroup et al\mbox.\egroup
  }{2018}]{li2018spatio}
Li, C.; Cui, Z.; Zheng, W.; Xu, C.; and Yang, J.
\newblock 2018.
\newblock Spatio-temporal graph convolution for skeleton based action
  recognition.
\newblock In {\em Thirty-Second AAAI Conference on Artificial Intelligence}.

\bibitem[\protect\citeauthoryear{Li \bgroup et al\mbox.\egroup
  }{2019a}]{li2019spatio}
Li, B.; Li, X.; Zhang, Z.; and Wu, F.
\newblock 2019a.
\newblock Spatio-temporal graph routing for skeleton-based action recognition.

\bibitem[\protect\citeauthoryear{Li \bgroup et al\mbox.\egroup
  }{2019b}]{li2019actional}
Li, M.; Chen, S.; Chen, X.; Zhang, Y.; Wang, Y.; and Tian, Q.
\newblock 2019b.
\newblock Actional-structural graph convolutional networks for skeleton-based
  action recognition.
\newblock In {\em IEEE Conference on Computer Vision and Pattern Recognition}.

\bibitem[\protect\citeauthoryear{Liu and Yuan}{2018}]{liu2018recognizing}
Liu, M., and Yuan, J.
\newblock 2018.
\newblock Recognizing human actions as the evolution of pose estimation maps.
\newblock In {\em Proceedings of the IEEE Conference on Computer Vision and
  Pattern Recognition},  1159--1168.

\bibitem[\protect\citeauthoryear{Liu \bgroup et al\mbox.\egroup
  }{2016}]{liu2016spatio}
Liu, J.; Shahroudy, A.; Xu, D.; and Wang, G.
\newblock 2016.
\newblock Spatio-temporal lstm with trust gates for 3d human action
  recognition.
\newblock In {\em European Conference on Computer Vision},  816--833.
\newblock Springer.

\bibitem[\protect\citeauthoryear{Liu \bgroup et al\mbox.\egroup
  }{2017a}]{liu2017skeleton}
Liu, J.; Shahroudy, A.; Xu, D.; Kot, A.~C.; and Wang, G.
\newblock 2017a.
\newblock Skeleton-based action recognition using spatio-temporal lstm network
  with trust gates.
\newblock {\em IEEE transactions on pattern analysis and machine intelligence}
  40(12):3007--3021.

\bibitem[\protect\citeauthoryear{Liu \bgroup et al\mbox.\egroup
  }{2017b}]{liu2017skeleton2}
Liu, J.; Wang, G.; Duan, L.-Y.; Abdiyeva, K.; and Kot, A.~C.
\newblock 2017b.
\newblock Skeleton-based human action recognition with global context-aware
  attention lstm networks.
\newblock {\em IEEE Transactions on Image Processing} 27(4):1586--1599.

\bibitem[\protect\citeauthoryear{Liu \bgroup et al\mbox.\egroup
  }{2017c}]{liu2017global}
Liu, J.; Wang, G.; Hu, P.; Duan, L.-Y.; and Kot, A.~C.
\newblock 2017c.
\newblock Global context-aware attention lstm networks for 3d action
  recognition.
\newblock In {\em Proceedings of the IEEE Conference on Computer Vision and
  Pattern Recognition},  1647--1656.

\bibitem[\protect\citeauthoryear{Liu \bgroup et al\mbox.\egroup
  }{2019a}]{liu2019ntu}
Liu, J.; Shahroudy, A.; Perez, M.~L.; Wang, G.; Duan, L.-Y.; and Chichung,
  A.~K.
\newblock 2019a.
\newblock Ntu rgb+ d 120: A large-scale benchmark for 3d human activity
  understanding.
\newblock {\em IEEE transactions on pattern analysis and machine intelligence}.

\bibitem[\protect\citeauthoryear{Liu \bgroup et al\mbox.\egroup
  }{2019b}]{liu2019skeleton}
Liu, J.; Shahroudy, A.; Wang, G.; Duan, L.-Y.; and Chichung, A.~K.
\newblock 2019b.
\newblock Skeleton-based online action prediction using scale selection
  network.
\newblock {\em IEEE transactions on pattern analysis and machine intelligence}.

\bibitem[\protect\citeauthoryear{Liu, Liu, and Chen}{2017}]{liu2017enhanced}
Liu, M.; Liu, H.; and Chen, C.
\newblock 2017.
\newblock Enhanced skeleton visualization for view invariant human action
  recognition.
\newblock {\em Pattern Recognition} 68:346--362.

\bibitem[\protect\citeauthoryear{Liu, Nickel, and
  Kiela}{2019}]{liu2019hyperbolic}
Liu, Q.; Nickel, M.; and Kiela, D.
\newblock 2019.
\newblock Hyperbolic graph neural networks.
\newblock In {\em Advances in Neural Information Processing Systems},
  8228--8239.

\bibitem[\protect\citeauthoryear{Mathieu \bgroup et al\mbox.\egroup
  }{2019}]{mathieu2019continuous}
Mathieu, E.; Le~Lan, C.; Maddison, C.~J.; Tomioka, R.; and Teh, Y.~W.
\newblock 2019.
\newblock Continuous hierarchical representations with poincar{\'e} variational
  auto-encoders.
\newblock In {\em Advances in neural information processing systems},
  12544--12555.

\bibitem[\protect\citeauthoryear{Monti \bgroup et al\mbox.\egroup
  }{2017}]{monti2017geometric}
Monti, F.; Boscaini, D.; Masci, J.; Rodola, E.; Svoboda, J.; and Bronstein,
  M.~M.
\newblock 2017.
\newblock Geometric deep learning on graphs and manifolds using mixture model
  cnns.
\newblock In {\em IEEE Conference on Computer Vision and Pattern Recognition},
  5115--5124.

\bibitem[\protect\citeauthoryear{Nickel and Kiela}{2017}]{nickel2017poincare}
Nickel, M., and Kiela, D.
\newblock 2017.
\newblock Poincar{\'e} embeddings for learning hierarchical representations.
\newblock In {\em Advances in neural information processing systems},
  6338--6347.

\bibitem[\protect\citeauthoryear{Peng \bgroup et al\mbox.\egroup
  }{2020}]{peng2020learning}
Peng, W.; Hong, X.; Chen, H.; and Zhao, G.
\newblock 2020.
\newblock Learning graph convolutional network for skeleton-based human action
  recognition by neural searching.
\newblock {\em Proceedings of the AAAI Conference on Artificial Intelligence}
  34(03):2669–2676.

\bibitem[\protect\citeauthoryear{Peng, Hong, and Zhao}{2019}]{peng2019video}
Peng, W.; Hong, X.; and Zhao, G.
\newblock 2019.
\newblock Video action recognition via neural architecture searching.
\newblock In {\em 2019 IEEE International Conference on Image Processing
  (ICIP)},  11--15.
\newblock IEEE.

\bibitem[\protect\citeauthoryear{Reynolds}{1993}]{reynolds1993hyperbolic}
Reynolds, W.~F.
\newblock 1993.
\newblock Hyperbolic geometry on a hyperboloid.
\newblock {\em The American mathematical monthly} 100(5):442--455.

\bibitem[\protect\citeauthoryear{Shahroudy \bgroup et al\mbox.\egroup
  }{2016}]{shahroudy2016ntu}
Shahroudy, A.; Liu, J.; Ng, T.-T.; and Wang, G.
\newblock 2016.
\newblock Ntu rgb+ d: A large scale dataset for 3d human activity analysis.
\newblock In {\em IEEE Conference on Computer Vision and Pattern Recognition},
  1010--1019.

\bibitem[\protect\citeauthoryear{Shi \bgroup et al\mbox.\egroup
  }{2019}]{shi2019two}
Shi, L.; Zhang, Y.; Cheng, J.; and Lu, H.
\newblock 2019.
\newblock Two-stream adaptive graph convolutional networks for skeleton-based
  action recognition.
\newblock In {\em IEEE Conference on Computer Vision and Pattern Recognition},
  12026--12035.

\bibitem[\protect\citeauthoryear{Si \bgroup et al\mbox.\egroup
  }{2018}]{si2018skeleton}
Si, C.; Jing, Y.; Wang, W.; Wang, L.; and Tan, T.
\newblock 2018.
\newblock Skeleton-based action recognition with spatial reasoning and temporal
  stack learning.
\newblock In {\em European Conference on Computer Vision},  103--118.

\bibitem[\protect\citeauthoryear{Song \bgroup et al\mbox.\egroup
  }{2017}]{song2017end}
Song, S.; Lan, C.; Xing, J.; Zeng, W.; and Liu, J.
\newblock 2017.
\newblock An end-to-end spatio-temporal attention model for human action
  recognition from skeleton data.
\newblock In {\em Thirty-first AAAI Conference on Artificial Intelligence}.

\bibitem[\protect\citeauthoryear{Tang \bgroup et al\mbox.\egroup
  }{2018}]{tang2018deep}
Tang, Y.; Tian, Y.; Lu, J.; Li, P.; and Zhou, J.
\newblock 2018.
\newblock Deep progressive reinforcement learning for skeleton-based action
  recognition.
\newblock In {\em IEEE Conference on Computer Vision and Pattern Recognition},
  5323--5332.

\bibitem[\protect\citeauthoryear{Tataru}{2001}]{tataru2001strichartz}
Tataru, D.
\newblock 2001.
\newblock Strichartz estimates in the hyperbolic space and global existence for
  the semilinear wave equation.
\newblock {\em Transactions of the American Mathematical society}
  353(2):795--807.

\bibitem[\protect\citeauthoryear{Tifrea, B{\'e}cigneul, and
  Ganea}{2018}]{tifrea2018poincar}
Tifrea, A.; B{\'e}cigneul, G.; and Ganea, O.-E.
\newblock 2018.
\newblock Poincar$\backslash$'e glove: Hyperbolic word embeddings.
\newblock {\em arXiv preprint arXiv:1810.06546}.

\bibitem[\protect\citeauthoryear{Veli{\v{c}}kovi{\'c} \bgroup et
  al\mbox.\egroup }{2018}]{velivckovic2018graph}
Veli{\v{c}}kovi{\'c}, P.; Cucurull, G.; Casanova, A.; Romero, A.; Lio, P.; and
  Bengio, Y.
\newblock 2018.
\newblock Graph attention networks.
\newblock {\em International Conference on Learning Representations}.

\bibitem[\protect\citeauthoryear{Vemulapalli, Arrate, and
  Chellappa}{2014}]{vemulapalli2014human}
Vemulapalli, R.; Arrate, F.; and Chellappa, R.
\newblock 2014.
\newblock Human action recognition by representing 3d skeletons as points in a
  lie group.
\newblock In {\em Proceedings of the IEEE conference on computer vision and
  pattern recognition},  588--595.

\bibitem[\protect\citeauthoryear{Wang \bgroup et al\mbox.\egroup
  }{2012}]{wang2012mining}
Wang, J.; Liu, Z.; Wu, Y.; and Yuan, J.
\newblock 2012.
\newblock Mining actionlet ensemble for action recognition with depth cameras.
\newblock In {\em 2012 IEEE Conference on Computer Vision and Pattern
  Recognition},  1290--1297.
\newblock IEEE.

\bibitem[\protect\citeauthoryear{Xia, Chen, and Aggarwal}{2012}]{xia2012view}
Xia, L.; Chen, C.-C.; and Aggarwal, J.~K.
\newblock 2012.
\newblock View invariant human action recognition using histograms of 3d
  joints.
\newblock In {\em IEEE Computer Society Conference on Computer Vision and
  Pattern Recognition Workshops},  20--27.
\newblock IEEE.

\bibitem[\protect\citeauthoryear{Yan, Xiong, and Lin}{2018}]{yan2018stgan}
Yan, S.; Xiong, Y.; and Lin, D.
\newblock 2018.
\newblock Spatial temporal graph convolutional networks for skeleton-based
  action recognition.
\newblock In {\em Thirty-Second AAAI Conference on Artificial Intelligence}.

\bibitem[\protect\citeauthoryear{Yu \bgroup et al\mbox.\egroup
  }{2018}]{yu2018slimmable}
Yu, J.; Yang, L.; Xu, N.; Yang, J.; and Huang, T.
\newblock 2018.
\newblock Slimmable neural networks.
\newblock {\em arXiv preprint arXiv:1812.08928}.

\bibitem[\protect\citeauthoryear{Zhang \bgroup et al\mbox.\egroup
  }{2017}]{zhang2017view}
Zhang, P.; Lan, C.; Xing, J.; Zeng, W.; Xue, J.; and Zheng, N.
\newblock 2017.
\newblock View adaptive recurrent neural networks for high performance human
  action recognition from skeleton data.
\newblock In {\em IEEE International Conference on Computer Vision},
  2117--2126.

\end{thebibliography}

\end{document}